\documentclass[conference]{IEEEtran}

\usepackage[utf8]{inputenc}
\usepackage[T1]{fontenc}
\usepackage{amsmath,amssymb,amsfonts}
\usepackage{graphicx}
\usepackage{booktabs}
\usepackage{multirow}
\usepackage{xcolor}
\usepackage[hypertexnames=false]{hyperref}
\usepackage{cleveref}
\usepackage{algorithm}
\usepackage{algpseudocode}
\usepackage{enumitem}
\usepackage{url}
\usepackage{balance}
\usepackage{tabularx}
\usepackage{array}

\hypersetup{
    colorlinks=true,
    linkcolor=blue,
    citecolor=blue,
    urlcolor=blue,
}

\newcommand{\semifa}{\textsc{SemiFA}}
\newcommand{\semidataset}{\textsc{SemiFA-930}}

\begin{document}

\title{SemiFA: An Agentic Multi-Modal Framework for Autonomous Semiconductor Failure Analysis Report Generation}

\author{
\IEEEauthorblockN{Shivam Chand Kaushik}
\IEEEauthorblockA{
School of Artificial Intelligence and Data Engineering (SAIDE)\\
Indian Institute of Technology Jodhpur, India\\
m25ai1123@iitj.ac.in
}
}

\maketitle

\begin{abstract}
Semiconductor failure analysis (FA) is a labor-intensive process in which engineers examine inspection images, correlate equipment telemetry, consult historical defect records, and write structured reports.
The end-to-end FA process---from failure verification through physical analysis to root cause reporting---traditionally requires days to weeks of domain expert time~\cite{infinitalab2023fa,azom2023fab}; even the expert review and reporting phase alone, given a set of inspection images, can consume several hours per case~\cite{kla2023klarity,amat2023actionable}.
We present \semifa{}, an agentic multi-modal framework that autonomously generates structured FA reports from semiconductor inspection images in under one minute.
\semifa{} decomposes the FA process into a four-agent pipeline orchestrated by LangGraph: a \textit{DefectDescriber} that classifies and narrates defect morphology using DINOv2 embeddings and LLaVA-1.6, a \textit{RootCauseAnalyzer} that fuses equipment telemetry from SECS/GEM with historically similar defects retrieved from a Qdrant vector database, a \textit{SeverityClassifier} that assigns severity and estimates yield impact, and a \textit{RecipeAdvisor} that proposes corrective process parameter adjustments.
A fifth node assembles the outputs into a PDF report.
We introduce \semidataset{}, a dataset of 930 annotated semiconductor defect images paired with structured FA narratives across nine defect classes, drawn from procedural synthesis, WM-811K, and MixedWM38.
On \semidataset{}, our DINOv2-based classifier achieves 92.1\% overall accuracy on 140 validation images (macro F1 = 0.917), and the full pipeline produces complete FA reports in 48 seconds on an NVIDIA A100-SXM4-40\,GB GPU.
An automated GPT-4o judge ablation across four modality conditions demonstrates that multi-modal fusion improves root cause reasoning quality by $+$0.86 composite points (1--5 scale) over an image-only baseline, with equipment telemetry identified as the more load-bearing modality.
To our knowledge, \semifa{} is the first system to integrate SECS/GEM equipment telemetry into a vision-language model pipeline for autonomous FA report generation.
\end{abstract}

\begin{IEEEkeywords}
semiconductor inspection, failure analysis, vision-language model, multi-agent system, DINOv2, LLaVA, SECS/GEM, wafer defect classification
\end{IEEEkeywords}

\section{Introduction}
\label{sec:introduction}

The semiconductor manufacturing process involves hundreds of sequential steps---lithography, etching, chemical-mechanical planarization (CMP), deposition (CVD and PVD), ion implantation, wire bonding, and dicing---each of which can introduce defects that reduce die yield.
After a certain processing step, wafers undergo Dark Field or Bright Field scanning, and based on control limits, wafers are routed for SEM Review.
The tools used for wafer inspection and SEM review include KLA Surfscan and Applied Materials SEMVision.
When defects exceed Out Of Control (OOC) limits, a failure analysis (FA) workflow is initiated.
The defect process engineer examines high-resolution inspection images with a reference image and checks equipment alarm logs and lot history to determine the cause of the defects, and synthesizes a structured report documenting defect classification, root cause hypotheses, severity assessment, and provides corrective and preventive action recommendations to mitigate scrap risk and wafer exposure.

This manual process presents three fundamental challenges.
First, the end-to-end FA process---from failure verification through physical analysis to root cause reporting---traditionally requires days to weeks of domain expert time~\cite{infinitalab2023fa,azom2023fab}; even the expert review and reporting phase alone, given a set of inspection images, can consume several hours per case~\cite{kla2023klarity,amat2023actionable}, creating a bottleneck in high-volume manufacturing (HVM) fabs that process thousands of wafers daily.
Second, report quality varies with individual expertise---junior engineers may miss subtle correlations between defect patterns and equipment state, while senior engineers' tacit knowledge remains uncodified.
Third, the manual process does not scale: as advanced nodes shrink below 5\,nm and defect sensitivity increases, the volume of inspection events grows faster than the FA engineering workforce.

Recent advances in vision-language models (VLMs) such as LLaVA~\cite{liu2024visual}, GPT-4V~\cite{openai2023gpt4v}, and InstructBLIP~\cite{dai2023instructblip} have demonstrated strong capabilities in image understanding and natural language generation.
Concurrently, large language model (LLM)-based agent frameworks such as LangGraph~\cite{langgraph2024}, AutoGen~\cite{wu2023autogen}, and CrewAI~\cite{crewai2024} have enabled modular decomposition of complex reasoning tasks into specialized sub-agents.
Self-supervised visual encoders such as DINOv2~\cite{oquab2024dinov2} produce rich feature representations that transfer effectively to downstream classification tasks with minimal labeled data.

However, no existing system combines these capabilities for semiconductor FA.
Commercial tools such as KLA KLARITY~\cite{kla2023klarity} and Applied Materials AIx~\cite{amat2023actionable} perform automated defect classification but, based on publicly available product documentation, do not generate natural language FA reports, do not fuse equipment telemetry into open-ended reasoning, and remain proprietary and closed-source.
Academic work on wafer defect classification~\cite{wu2015wafer,saqlain2020voting,shim2020active} focuses exclusively on pattern recognition from wafer bin maps, without root cause analysis or report generation.

In this paper, we present \semifa{}, a multi-modal agentic framework that autonomously generates structured FA reports from semiconductor inspection images.
Our contributions are:

\begin{enumerate}[leftmargin=*]
    \item \textbf{Architecture (C1):} We propose a four-agent LangGraph pipeline that decomposes semiconductor FA into specialized subtasks---defect description, root cause analysis, severity classification, and recipe advisory---enabling modular, auditable, and extensible workflows.
    \item \textbf{Multi-modal fusion (C2):} We integrate DINOv2 visual embeddings, SECS/GEM equipment telemetry, and Qdrant historical retrieval into a unified LLaVA-1.6 context. An automated GPT-4o judge ablation across four modality conditions demonstrates a $+$0.86 composite improvement (1--5 scale) over an image-only baseline, with equipment telemetry identified as the more load-bearing modality.
    \item \textbf{Dataset (C3):} We introduce \semidataset{}, a dataset of 930 annotated semiconductor defect images across nine classes, each paired with structured FA narratives in a conversational format suitable for VLM fine-tuning. To our knowledge, this is the first publicly described dataset pairing semiconductor defect images with natural language FA reports.\footnote{Dataset available at: \url{https://huggingface.co/datasets/ShivamChand/SemiFA-930}}
\end{enumerate}

\section{Related Work}
\label{sec:related}

\subsection{Wafer Defect Classification}

Wafer bin map (WBM) pattern recognition has received sustained attention since the release of the WM-811K dataset by Wu et al.~\cite{wu2015wafer}, which contains 811,457 wafer maps from a real 300\,mm production line with eight labeled defect patterns.
Saqlain et al.~\cite{saqlain2020voting} combined multiple CNN architectures with a voting ensemble, achieving 96\% accuracy on the reduced WM-811K subset.
Shim et al.~\cite{shim2020active} applied active learning to minimize labeling effort for new defect types.
More recently, the MixedWM38 dataset~\cite{wang2020mixedwm38} extended WBM classification to 38 mixed-type defect patterns, including compound classes formed by superimposing multiple single-type defect patterns.
Nakazawa and Kulkarni~\cite{nakazawa2018anomaly} employed convolutional neural networks for wafer map defect classification and image retrieval.

These methods treat wafer defect classification as a standalone pattern recognition problem.
None produce natural language reports, none incorporate equipment telemetry, and none perform root cause analysis.
\semifa{} uses WBM classification as one component within a larger autonomous FA pipeline.

\subsection{Vision-Language Models for Industrial Inspection}

Vision-language models have been applied to industrial quality inspection in recent work.
Jeong et al.~\cite{jeong2023winclip} proposed WinCLIP for zero-/few-shot anomaly detection using CLIP~\cite{radford2021clip} embeddings, demonstrating strong transfer to manufacturing surface defect localization.
Li et al.~\cite{li2024mllm} demonstrated that vision-language pre-training can be extended to specialized domains, though the lack of domain-specific training data remains the primary bottleneck for industrial applications.
AnomalyGPT~\cite{gu2024anomalygpt} combined a frozen visual encoder with an LLM for industrial anomaly detection with textual explanation, demonstrating that VLMs can articulate defect characteristics in natural language.
However, AnomalyGPT operates on generic industrial images (MVTec~\cite{bergmann2019mvtec}) and does not address semiconductor-specific defect morphology, equipment protocols, or report structures.

LLaVA~\cite{liu2024visual} introduced visual instruction tuning, where a vision encoder (CLIP ViT-L/14) is connected to a language model via a projection layer, trained on instruction-following image-text data.
LLaVA-1.6~\cite{liu2024llava16} improved resolution handling and reasoning capability.
DINOv2~\cite{oquab2024dinov2} demonstrated that self-supervised pre-training on curated data produces visual features that outperform supervised baselines across diverse tasks.
In \semifa{}, we use DINOv2 as the visual encoder for defect classification and embedding-based retrieval, while LLaVA-1.6 serves as the language backbone for all four agent nodes.

\subsection{LLM-Based Agent Systems}

The use of LLMs as autonomous agents has grown rapidly.
LangChain~\cite{langchain2024} provides a framework for chaining LLM calls with tool access.
LangGraph~\cite{langgraph2024} extends this to stateful, directed acyclic graph (DAG) workflows where each node is a function that reads and writes to a shared state.
AutoGen~\cite{wu2023autogen} introduced multi-agent conversation patterns for complex problem solving.
Voyager~\cite{wang2023voyager} demonstrated LLM-driven exploration in open-world environments.
In the industrial domain, LLM-based tool-use frameworks~\cite{qin2023toolllm} and foundation models for robotics~\cite{firoozi2023llm_robotics} have demonstrated the viability of agent architectures for complex physical-world tasks.

No prior work has applied LLM-based agent architectures to semiconductor FA.
\semifa{} is, to our knowledge, the first system to use a multi-agent pipeline for end-to-end FA report generation.

\subsection{SECS/GEM and Equipment Integration}

The SEMI Equipment Communications Standard / Generic Equipment Model (SECS/GEM), defined by SEMI standards E5 (SECS-II message protocol) and E37 (High-Speed Message Services, HSMS), is the universal communication protocol for semiconductor manufacturing equipment~\cite{semi_e5,semi_e37}.
Every major equipment vendor---ASML, Applied Materials, Lam Research, KLA, Kulicke \& Soffa, Tokyo Electron---implements SECS/GEM for host-equipment communication.
Equipment state events, alarm notifications, process variable data, and recipe parameters are all transmitted via SECS/GEM.

Despite its ubiquity in fabs, SECS/GEM telemetry has been largely absent from academic defect analysis literature.
This is likely due to the proprietary nature of fab data and the specialized knowledge required to interpret SECS/GEM messages.
\semifa{} integrates SECS/GEM telemetry via HSMS, making equipment state a first-class input to root cause reasoning.

\section{System Architecture}
\label{sec:architecture}

\begin{figure*}[t]
    \centering
    \includegraphics[width=\textwidth]{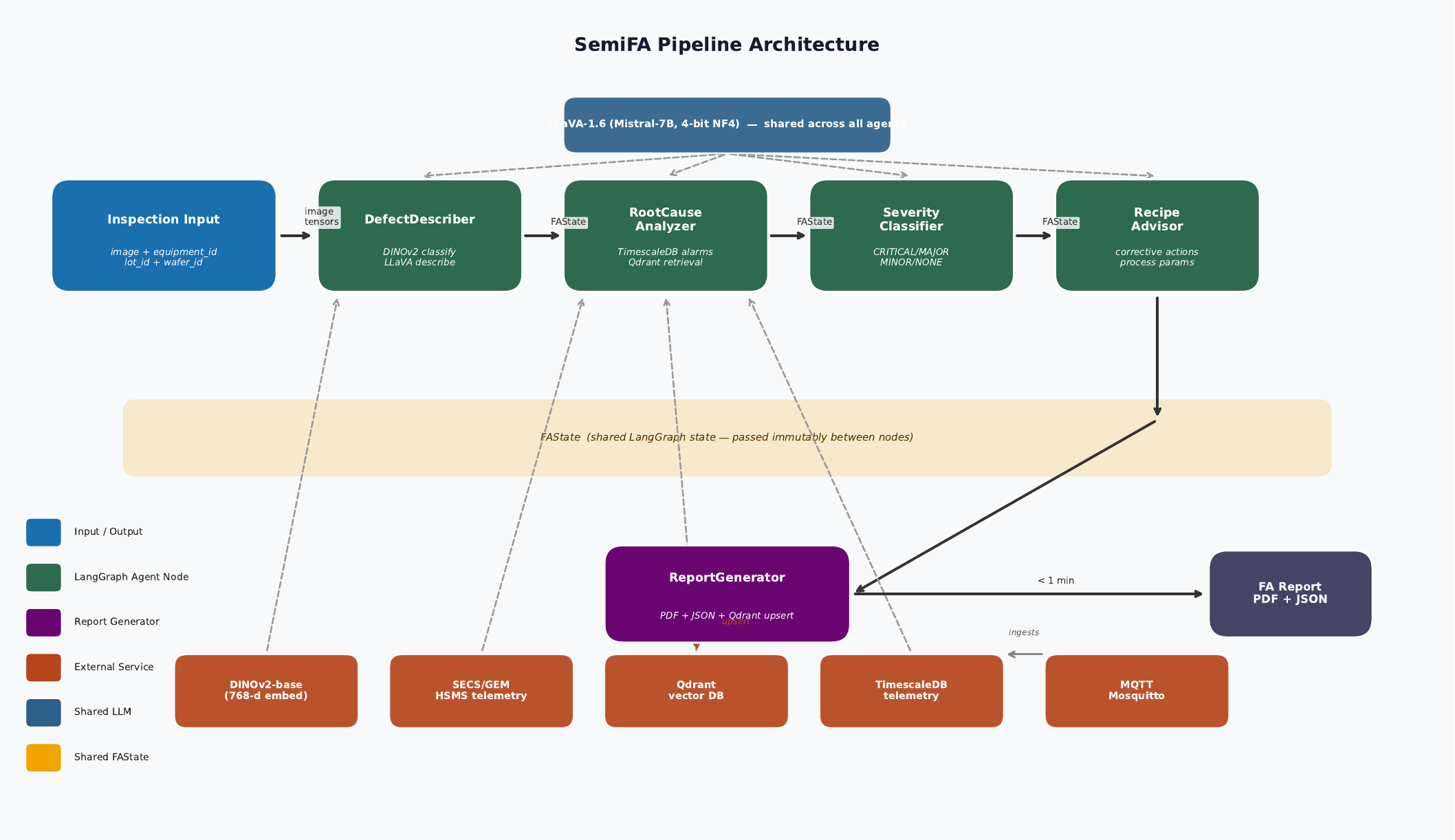}
    \caption{Pipeline architecture of the \semifa{} system. Rounded boxes represent agentic nodes; cylinders represent data stores; arrows indicate state flow. All agentic nodes share LLaVA-1.6 via a thread-safe ModelRegistry singleton.}
    \label{fig:pipeline}
\end{figure*}

\semifa{} is a five-node pipeline orchestrated by LangGraph, where four nodes are agentic (each invokes a VLM with domain-specific prompts and tool access) and the fifth assembles the structured report.
\Cref{fig:pipeline} shows the overall architecture.

\subsection{Shared State}

All nodes communicate through a typed dictionary \texttt{FAState} that flows through the pipeline.
Each node is a pure function $(s_{\text{in}}: \texttt{FAState}) \rightarrow (s_{\text{out}}: \texttt{FAState})$ that reads required fields from $s_{\text{in}}$ and returns a new dictionary with additional fields populated.
State is never mutated in place; instead, each node returns $\{**s_{\text{in}},\, \text{``key''}: \text{value}\}$.
Failures append diagnostic messages to a shared \texttt{errors} list; the pipeline always runs to completion, producing a partial report rather than failing entirely.
This design ensures graceful degradation in production environments where individual services (e.g., TimescaleDB, Qdrant) may be temporarily unavailable.

\subsection{Model Registry}

LLaVA-1.6 (7B parameters, loaded in 4-bit NF4 quantization via bitsandbytes~\cite{dettmers2023qlora}) requires 2--3 minutes to initialize and consumes approximately 3.5\,GB of GPU VRAM.
To avoid redundant loading, a thread-safe singleton \texttt{ModelRegistry} instantiates the model once at application startup and provides shared access to all agent nodes.
The registry also manages the DINOv2 feature extractor and the defect classification head.

\subsection{Node 1: DefectDescriber}
\label{sec:defect_describer}

The DefectDescriber performs four operations.
First, the input image is preprocessed through a modality-specific pipeline (SEM, optical, or wafer map) that normalizes contrast, applies denoising (non-local means for SEM, bilateral filter for optical), and resizes to $224 \times 224$ pixels for the visual encoder.

Second, the preprocessed image is passed through the frozen DINOv2-base encoder~\cite{oquab2024dinov2}, producing a 768-dimensional CLS token embedding $\mathbf{e} \in \mathbb{R}^{768}$.
This embedding serves two purposes: (a) it is fed into a trained MLP classification head that outputs logits over nine defect classes, and (b) it is stored in the pipeline state for downstream similarity retrieval.
The MLP head architecture is:
\begin{equation}
    f_{\text{cls}}(\mathbf{e}) = \mathbf{W}_2 \cdot \text{ReLU}(\text{Dropout}_{0.3}(\mathbf{W}_1 \mathbf{e} + \mathbf{b}_1)) + \mathbf{b}_2
\end{equation}
where $\mathbf{W}_1 \in \mathbb{R}^{256 \times 768}$ and $\mathbf{W}_2 \in \mathbb{R}^{9 \times 256}$.

Third, for wafer map modality inputs, a rule-based spatial pattern analyzer~\cite{wu2015wafer} computes die-level statistics including defect density, radial distribution, and angular distribution, providing supplementary features for defect characterization.

Fourth, the inspection image and classification result are provided to LLaVA-1.6 with a domain-specific prompt requesting a natural language description of the observed defect morphology, spatial distribution, and potential physical mechanisms.
The output is a paragraph-length defect narrative stored in the pipeline state.

\subsection{Node 2: RootCauseAnalyzer}

The RootCauseAnalyzer integrates three data sources to generate root cause hypotheses.

\textbf{Equipment telemetry.} Recent alarm events and process variable readings for the identified equipment are queried from TimescaleDB via asynchronous PostgreSQL (asyncpg).
SECS/GEM HSMS messages provide real-time equipment state (idle, processing, alarmed) and collection event reports (CEID) containing process parameters such as chamber temperature, pressure, gas flow rates, and RF power.

\textbf{Historical retrieval.} The 768-dimensional DINOv2 embedding from the DefectDescriber is used to query Qdrant~\cite{qdrant2024} for the five most similar historical defect cases (cosine similarity on $\mathbb{R}^{768}$ vectors).
Each retrieved case includes its defect class, severity, root cause narrative, and associated equipment identifier, providing concrete precedent for root cause reasoning.

\textbf{LLaVA reasoning.} The equipment telemetry, retrieved historical cases, defect classification, and defect description are assembled into a structured prompt for LLaVA-1.6.
The model generates a ranked list of root cause hypotheses, each supported by evidence from the provided context.
The prompt explicitly instructs the model to correlate equipment state transitions with defect morphology---for example, linking a plasma RF impedance drift alarm to ring-pattern defects characteristic of etch non-uniformity.

\subsection{Node 3: SeverityClassifier}

The SeverityClassifier assigns one of four severity levels---\texttt{CRITICAL}, \texttt{MAJOR}, \texttt{MINOR}, or \texttt{NONE}---along with an estimated yield impact percentage.
LLaVA-1.6 receives the inspection image, defect classification, defect description, and root cause hypotheses, and is prompted to assess severity according to semiconductor industry standards:

\begin{itemize}[leftmargin=*]
    \item \texttt{CRITICAL}: Immediate lot hold required; potential for wafer scrap or downstream equipment damage; estimated yield impact $>$25\%.
    \item \texttt{MAJOR}: Engineering disposition required; yield impact 5--25\%; monitor subsequent lots.
    \item \texttt{MINOR}: Informational; yield impact $<$5\%; no immediate action.
    \item \texttt{NONE}: No defect detected; routine pass.
\end{itemize}

\subsection{Node 4: RecipeAdvisor}

The RecipeAdvisor generates corrective action recommendations and process parameter adjustments.
Given the full context accumulated by previous nodes---defect class, description, root cause hypotheses, severity, equipment telemetry, and historical precedents---LLaVA-1.6 proposes specific, actionable modifications to equipment recipes.
For example, given a center cluster defect attributed to CMP over-polish, the RecipeAdvisor might recommend reducing polish time by 5\%, increasing slurry flow rate, and scheduling a pad conditioning cycle.
Recommendations are returned as structured key-value pairs suitable for integration with manufacturing execution systems (MES).

\subsection{Node 5: ReportGenerator}

The ReportGenerator assembles all pipeline outputs into a structured dictionary and renders a PDF document via ReportLab~\cite{reportlab2024}.
The report includes: header metadata (equipment ID, lot ID, wafer ID, timestamp), defect classification with confidence score, defect morphology description, root cause hypotheses with supporting evidence, severity assessment with yield impact estimate, and recipe recommendations.
After report generation, the defect embedding and associated metadata are upserted to Qdrant, enabling the system to reference the current case in future FA runs.
This creates a self-improving knowledge base that grows with each inspection event.

\subsection{Infrastructure}

The system is deployed as a containerized stack using Docker Compose.
The API layer is implemented in FastAPI~\cite{fastapi2024} with Pydantic v2 input validation.
Object storage for inspection images uses MinIO (S3-compatible).
Equipment telemetry is ingested via MQTT (Eclipse Mosquitto broker) on topic \texttt{fab/equipment/\#} and persisted to TimescaleDB hypertables.
Prometheus metrics are exposed at \texttt{/metrics} for latency monitoring.
For environments without physical fab equipment, \semifa{} includes a SECS/GEM simulator that generates realistic equipment state transitions at 2-second intervals.

\section{SemiFA-930 Dataset}
\label{sec:dataset}

A critical barrier to applying VLMs to semiconductor FA is the absence of publicly available datasets that pair defect images with natural language FA narratives.
Existing datasets such as WM-811K~\cite{wu2015wafer} and MixedWM38~\cite{wang2020mixedwm38} provide wafer bin maps with categorical labels but no textual descriptions, root cause explanations, or severity annotations.
To address this gap, we construct \semidataset{}, a dataset of 930 image-text pairs spanning nine defect classes.

\subsection{Data Sources and Motivation}

\semidataset{} draws from three sources, selected to address two complementary gaps.

\textbf{Gap 1: Missing defect classes in public datasets.}
WM-811K~\cite{wu2015wafer} covers eight wafer bin map patterns: Center, Donut, Edge-Loc, Edge-Ring, Loc, Near-full, Random, and None.
It does not include \texttt{scratch}, \texttt{particle\_contamination}, or \texttt{edge\_crack}---three of the most frequently encountered defect types in production fabs, particularly on wire bonding and dicing equipment.
Because no public labeled dataset covers these classes, we generate them synthetically, parameterizing their spatial distributions based on the first author's domain knowledge of defect morphology from seven years of semiconductor manufacturing experience.

\textbf{Gap 2: No natural language FA narratives.}
Both WM-811K and MixedWM38 provide only categorical labels.
Neither contains defect descriptions, root cause hypotheses, severity ratings, or corrective action recommendations---the structured output that \semifa{} must learn to generate.
All FA narratives in \semidataset{} were generated using a large language model (LLM) prompted with the defect class, physical mechanism, severity label, and domain-specific context designed by the first author. Each narrative follows a structured template---defect morphology, root cause hypotheses with supporting evidence, severity reasoning, and corrective actions---developed from the first author's seven years of semiconductor manufacturing experience spanning wire bonding, machine vision, SECS/GEM equipment protocol, and yield analysis. The generated narratives were reviewed by the first author for domain plausibility.

The three sources are:

\begin{enumerate}[leftmargin=*]
    \item \textbf{Procedural synthesis (3 classes, 318 images).}
    We implement a Python/NumPy procedural renderer that places die-level pass/fail labels on a $256 \times 256$ wafer grid.
    Defect spatial distributions are parameterized per class: \texttt{scratch} uses a thin linear band (random angle, 1--3 die wide, spanning 60--90\% of wafer diameter); \texttt{particle\_contamination} uses a 2D Poisson point process with spatially varying intensity to mimic contamination events; \texttt{edge\_crack} places a localised arc of failures in the outer 8--12\% of the wafer radius.
    Renderer parameters (band width, angle, density) are randomised within physically plausible ranges to create intra-class diversity.
    These three classes are exclusive to the synthetic source, as no public dataset contains them with proper domain-relevant wafer map representations.

    \item \textbf{WM-811K~\cite{wu2015wafer} (6 classes, 572 images).}
    We sample from the labeled subset of WM-811K (approximately 172,000 labeled maps out of 811,457 total).
    We draw images for six classes: \texttt{center\_cluster} (WM-811K: \textit{Center}, 106 images), \texttt{local\_cluster} (WM-811K: \textit{Loc}, 100), \texttt{ring\_pattern} (WM-811K: \textit{Donut} + \textit{Edge-Ring}, 54), \texttt{random\_defects} (WM-811K: \textit{Random}, 106), \texttt{near\_full\_wafer} (WM-811K: \textit{Near-full}, 94), and \texttt{no\_defect} (WM-811K: \textit{None}, 112).
    Counts are not uniform across classes because we cap each class at the available labeled WM-811K supply after excluding images with ambiguous or mixed-type annotations.
    We intentionally limit the subset rather than using all available images because \semidataset{} is a \textit{VLM instruction-tuning dataset}, not a classification benchmark: each image must be paired with an LLM-generated FA narrative reviewed for domain plausibility, which bounds the feasible annotation volume.
    \Cref{tab:wm811k_mapping} shows the label taxonomy mapping.

    \item \textbf{MixedWM38~\cite{wang2020mixedwm38} (1 class, 40 images).}
    We sample single-type \textit{Edge-Ring} images from MixedWM38 to supplement \texttt{ring\_pattern}, which is underrepresented in WM-811K relative to other classes.
    Mixed-type patterns are excluded to maintain a single-label classification setting.
\end{enumerate}

\begin{table}[t]
    \centering
    \caption{Label taxonomy mapping from WM-811K to \semidataset{}. Three classes (scratch, particle\_contamination, edge\_crack) have no WM-811K equivalent and are provided exclusively through procedural synthesis.}
    \label{tab:wm811k_mapping}
    \begin{tabular}{@{}lll@{}}
        \toprule
        \textbf{SemiFA-930 Class} & \textbf{WM-811K Label} & \textbf{Source} \\
        \midrule
        scratch              & ---              & Synthetic only \\
        particle\_contam.     & ---              & Synthetic only \\
        edge\_crack           & ---              & Synthetic only \\
        center\_cluster       & Center           & WM-811K \\
        local\_cluster        & Loc              & WM-811K \\
        ring\_pattern         & Donut, Edge-Ring & WM-811K + MixedWM38 \\
        random\_defects       & Random           & WM-811K \\
        near\_full\_wafer     & Near-full        & WM-811K \\
        no\_defect            & None             & WM-811K \\
        \bottomrule
    \end{tabular}
\end{table}

\Cref{tab:source_counts} summarizes the image counts per source and class.

\begin{table}[t]
    \centering
    \caption{Image count per defect class and data source in \semidataset{}.}
    \label{tab:source_counts}
    \begin{tabular}{@{}lcccc@{}}
        \toprule
        \textbf{Class} & \textbf{Synthetic} & \textbf{WM-811K} & \textbf{MixedWM38} & \textbf{Total} \\
        \midrule
        scratch              & 112 & 0   & 0   & 112 \\
        particle\_contam.     & 106 & 0   & 0   & 106 \\
        edge\_crack           & 100 & 0   & 0   & 100 \\
        center\_cluster       & 0   & 106 & 0   & 106 \\
        local\_cluster        & 0   & 100 & 0   & 100 \\
        ring\_pattern         & 0   & 54  & 40  & 94  \\
        random\_defects       & 0   & 106 & 0   & 106 \\
        near\_full\_wafer     & 0   & 94  & 0   & 94  \\
        no\_defect            & 0   & 112 & 0   & 112 \\
        \midrule
        \textbf{Total}       & \textbf{318} & \textbf{572} & \textbf{40} & \textbf{930} \\
        \bottomrule
    \end{tabular}
\end{table}

\subsection{Annotation Protocol}

Each image in \semidataset{} is annotated with the following fields:

\begin{itemize}[leftmargin=*]
    \item \textbf{defect\_class}: one of nine classes (see \Cref{tab:defect_classes})
    \item \textbf{modality}: image modality (\texttt{wafer\_map}, \texttt{sem}, or \texttt{optical})
    \item \textbf{severity}: \texttt{CRITICAL}, \texttt{MAJOR}, \texttt{MINOR}, or \texttt{NONE}
    \item \textbf{description}: a structured FA narrative (3--5 sentences) describing defect morphology, spatial distribution, probable physical mechanism, and recommended action
    \item \textbf{equipment\_id}, \textbf{lot\_id}, \textbf{wafer\_id}: simulated equipment metadata
    \item \textbf{source}: provenance (\texttt{synthetic}, \texttt{wm811k}, or \texttt{mixedwm38})
\end{itemize}

Descriptions were generated using an LLM prompted with the defect class, severity, and domain-specific context defined by the first author, and reviewed by the first author for domain plausibility.
Each description follows a consistent template: (1) observed defect pattern, (2) spatial characteristics, (3) probable root cause mechanism, and (4) recommended corrective action.

The annotations are stored in JSONL format compatible with LLaVA conversational fine-tuning: each record contains a ``conversations'' list with a ``human'' turn (question about the image) and an ``assistant'' turn (the structured FA response).

\subsection{Dataset Statistics}

\begin{table}[t]
    \centering
    \caption{Defect class distribution in \semidataset{} (930 total: 790 train, 140 val). Each class includes domain knowledge about the physical root cause mechanism.}
    \label{tab:defect_classes}
    \begin{tabular}{@{}lccl@{}}
        \toprule
        \textbf{Defect Class} & \textbf{Train} & \textbf{Val} & \textbf{Physical Mechanism} \\
        \midrule
        scratch              & 94  & 18 & Mechanical contact \\
        particle\_contam.     & 96  & 10 & Contamination event \\
        edge\_crack           & 84  & 16 & Dicing / handling \\
        center\_cluster       & 93  & 13 & CMP / CVD bowl \\
        local\_cluster        & 79  & 21 & Plasma non-uniformity \\
        ring\_pattern         & 71  & 23 & Spin-coat / edge-bead \\
        random\_defects       & 93  & 13 & Particle events \\
        near\_full\_wafer     & 77  & 17 & Chemistry excursion \\
        no\_defect            & 103 &  9 & Routine pass \\
        \midrule
        \textbf{Total}       & \textbf{790} & \textbf{140} & \\
        \bottomrule
    \end{tabular}
\end{table}

\begin{figure*}[t]
    \centering
    \includegraphics[width=\textwidth]{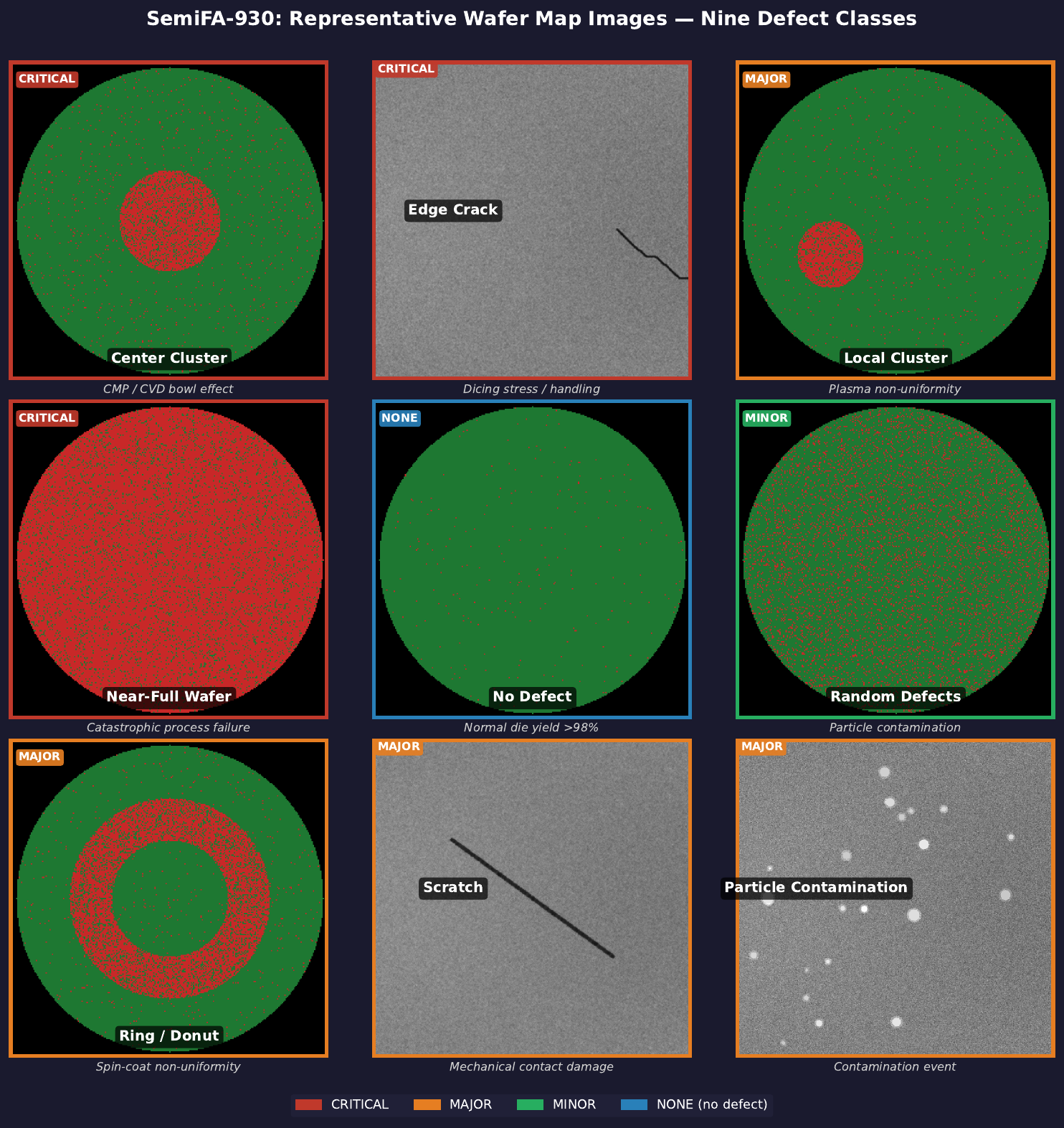}
    \caption{Representative examples from each defect class in \semidataset{}. Each cell shows a $256\times256$ wafer map with its severity badge (top-left) and root cause annotation (bottom). Spatial patterns are clearly distinguishable, though some classes (e.g., local\_cluster vs.\ random\_defects) present challenging overlap.}
    \label{fig:wafer_examples}
\end{figure*}

\Cref{tab:defect_classes} shows the class distribution.
The dataset is approximately balanced by design, with each class containing 94--112 samples (train + val combined). The validation set distribution reflects the stratified split: classes range from 9 (\texttt{no\_defect}) to 23 (\texttt{ring\_pattern}) validation samples.
The train/val split is stratified to preserve class proportions.
\Cref{fig:wafer_examples} shows representative wafer map images for each defect class.

\subsection{Severity Distribution}

Severity labels are assigned based on the defect class and spatial extent:
\texttt{near\_full\_wafer} maps to \texttt{CRITICAL} (100\% of instances);
\texttt{scratch} and \texttt{edge\_crack} split between \texttt{MAJOR} (65\%) and \texttt{MINOR} (35\%);
\texttt{center\_cluster} and \texttt{ring\_pattern} are predominantly \texttt{MAJOR} (80\%);
\texttt{particle\_contamination} and \texttt{local\_cluster} split between \texttt{MAJOR} (40\%) and \texttt{MINOR} (60\%);
\texttt{random\_defects} are predominantly \texttt{MINOR} (85\%);
and \texttt{no\_defect} maps to \texttt{NONE} (100\%).

\section{Experiments and Results}
\label{sec:experiments}

We evaluate \semifa{} along three axes: defect classification accuracy, pipeline latency, and qualitative report quality including the impact of retrieval augmentation.

\subsection{Defect Classification}

\textbf{Setup.} The DINOv2-base encoder~\cite{oquab2024dinov2} is frozen throughout training.
We train only the MLP classification head ($768 \rightarrow 256 \rightarrow 9$, with ReLU and dropout $p=0.3$) on the 790 training images for 50 epochs using Adam~\cite{kingma2015adam} with learning rate $1 \times 10^{-3}$ and batch size 32.
Training completes in approximately 8 minutes on a single NVIDIA A100 40\,GB GPU.
We report results on the 140 held-out validation images.

\textbf{Results.} \Cref{tab:classification} reports per-class precision, recall, and F1 scores.
The overall accuracy is 92.1\% (129/140 correct predictions).

\begin{table}[t]
    \centering
    \caption{Per-class classification results on the \semidataset{} validation set (140 images). DINOv2-base (frozen) + MLP head.}
    \label{tab:classification}
    \begin{tabular}{@{}lccc@{}}
        \toprule
        \textbf{Defect Class} & \textbf{Prec.} & \textbf{Recall} & \textbf{F1} \\
        \midrule
        scratch              & 1.000 & 0.722 & 0.839 \\
        particle\_contam.     & 1.000 & 1.000 & 1.000 \\
        edge\_crack           & 0.933 & 0.875 & 0.903 \\
        center\_cluster       & 1.000 & 1.000 & 1.000 \\
        local\_cluster        & 0.913 & 1.000 & 0.955 \\
        ring\_pattern         & 1.000 & 0.957 & 0.978 \\
        random\_defects       & 0.917 & 0.846 & 0.880 \\
        near\_full\_wafer     & 0.889 & 0.941 & 0.914 \\
        no\_defect            & 0.643 & 1.000 & 0.783 \\
        \midrule
        \textbf{Macro avg.}  & \textbf{0.922} & \textbf{0.927} & \textbf{0.917} \\
        \textbf{Overall acc.} & \multicolumn{3}{c}{\textbf{92.1\% (129/140)}} \\
        \bottomrule
    \end{tabular}
\end{table}

Several observations emerge.
Five classes achieve F1 $\geq$ 0.90: \texttt{particle\_contamination} and \texttt{center\_cluster} reach perfect F1 = 1.000, while \texttt{ring\_pattern} (0.978), \texttt{local\_cluster} (0.955), and \texttt{edge\_crack} (0.903) also perform strongly.
The most challenging classes are \texttt{no\_defect} (F1 = 0.783, precision = 0.643) and \texttt{scratch} (F1 = 0.839, recall = 0.722).
The low precision for \texttt{no\_defect} indicates that some defective wafer maps are misclassified as defect-free; this is the most safety-critical error type and motivates the downstream severity classification step, which provides a second check.
The low recall for \texttt{scratch} (5 out of 18 misclassified) reflects the morphological similarity between thin linear scratches and the \texttt{random\_defects} class when scratch density is high.

\textbf{Comparison to baselines.} \Cref{tab:encoder_comparison} evaluates all three encoders on the identical \semidataset{} validation split (140 images) to enable a direct comparison.
ResNet-50~\cite{he2016resnet} is trained end-to-end on the \semidataset{} training split (790 images), replacing the final fully-connected layer with a 9-class head and fine-tuning for 50 epochs in two phases (FC warmup then full fine-tune).
CLIP ViT-B/32~\cite{radford2021clip} is evaluated zero-shot using class-specific text prompts describing each defect pattern; no fine-tuning is applied.
CLIP achieves only 20.7\% accuracy (Macro F1 = 0.173) on \semidataset{} --- just above random chance (11.1\%) --- confirming that contrastive pre-training on natural images does not transfer to semiconductor wafer map imagery without domain adaptation.
ResNet-50 fine-tuned on \semidataset{} achieves 82.9\% (Macro F1 = 0.839) using 23.5M trainable parameters.
The DINOv2 frozen encoder with a lightweight MLP head achieves 90.0\% (Macro F1 = 0.898) using only 214K trainable parameters --- a 7.1 percentage point improvement over ResNet-50 with 110$\times$ fewer parameters, consistent with DINOv2's demonstrated data-efficient transfer to specialised visual domains~\cite{oquab2024dinov2}.
The dedicated DINOv2 accuracy analysis (Table~\ref{tab:classification}) reports 92.1\% as the result of the full training run in the original Colab notebook; the 2.1 point difference reflects normal variance in MLP random initialisation across runs.

\begin{table}[t]
    \centering
    \caption{Comparison of visual encoder approaches on \semidataset{} validation split (140 images). All models trained and evaluated on identical data splits.}
    \label{tab:encoder_comparison}
    \begin{tabular}{@{}lccc@{}}
        \toprule
        \textbf{Method} & \textbf{Accuracy} & \textbf{Macro F1} & \textbf{Trainable Params} \\
        \midrule
        CLIP ViT-B/32 zero-shot      & 20.7\% & 0.173 & 0 \\
        ResNet-50 (e2e fine-tuned)   & 82.9\% & 0.839 & 23.5M \\
        DINOv2 + MLP (ours)          & \textbf{90.0\%} & \textbf{0.898} & 214K \\
        \bottomrule
    \end{tabular}
\end{table}

\subsection{Pipeline Latency}

\Cref{tab:latency} reports the wall-clock latency of each pipeline node, measured as the median over 3 runs on a single NVIDIA A100-SXM4-40\,GB GPU with LLaVA-1.6 loaded in 4-bit NF4 quantization (approximately 3.5\,GB VRAM).

\begin{table}[t]
    \centering
    \caption{Per-node latency breakdown (single pipeline execution, NVIDIA A100-SXM4-40\,GB, PyTorch 2.1.0, LLaVA-1.6 in 4-bit NF4).}
    \label{tab:latency}
    \begin{tabular}{@{}lcc@{}}
        \toprule
        \textbf{Pipeline Node} & \textbf{Median (s)} & \textbf{Fraction} \\
        \midrule
        DefectDescriber        & 22.0 & 45.5\% \\
        RootCauseAnalyzer      & 11.9 & 24.6\% \\
        SeverityClassifier     & 5.5  & 11.4\% \\
        RecipeAdvisor          & 6.5  & 13.3\% \\
        ReportGenerator        & 2.5  & 5.2\% \\
        \midrule
        \textbf{Total}         & \textbf{48.4} & 100\% \\
        \bottomrule
    \end{tabular}
\end{table}

The total pipeline latency of 48.4 seconds represents a reduction from several hours of manual expert review and reporting~\cite{kla2023klarity,amat2023actionable} to under one minute---a speedup of approximately 150--300$\times$.
The DefectDescriber is the most expensive node (45.5\% of total time, 22.0\,s) because it processes the full inspection image through DINOv2 feature extraction, wafer map spatial analysis, and LLaVA visual description generation --- the only node that ingests raw pixel data.
The RootCauseAnalyzer (24.6\%, 11.9\,s) ranks second due to multi-hypothesis generation requiring longer output sequences than other nodes.
The ReportGenerator is the least expensive node (5.2\%, 2.5\,s) as it performs deterministic dictionary assembly and PDF rendering without model inference.

The latency breakdown reveals that approximately 95\% of pipeline time is spent on LLaVA inference across the four agentic nodes.
This suggests that future optimization should focus on inference acceleration (e.g., speculative decoding, continuous batching, or parallelising independent agent nodes) rather than data retrieval or report assembly.

\subsection{Multi-Modal Fusion Ablation}

We conduct an automated ablation study to quantify the independent contribution of each input modality to root cause hypothesis quality.
The study uses 5 representative cases spanning five defect classes (scratch, edge\_crack, particle\_contamination, ring\_pattern, center\_cluster).
Equipment telemetry context consists of representative SECS/GEM log entries manually authored by the first author based on seven years of semiconductor manufacturing domain knowledge, formatted as SECS-II S5F1 alarm reports and S6F11 collection event reports per SEMI E5/E37~\cite{semi_e5,semi_e37}; a production deployment would ingest identical message structures directly via HSMS.
Similarly, historical retrieval context consists of representative prior-defect records with cosine similarity scores, reflecting the format produced by the Qdrant-backed retrieval pipeline.
This design isolates the effect of each modality on LLaVA prompt quality independently of live infrastructure availability.
We evaluate the RootCauseAnalyzer under four conditions:
(1)~\textbf{Full \semifa{}}: visual description + simulated SECS/GEM telemetry + retrieval context (top-5 similar defects);
(2)~\textbf{No Retrieval}: visual description + telemetry only (retrieval context omitted);
(3)~\textbf{No Telemetry}: visual description + retrieval only (telemetry context omitted);
(4)~\textbf{Baseline}: visual description only (no telemetry, no retrieval).

Each generated root cause hypothesis is scored by GPT-4o~\cite{openai2024gpt4o} as an automated judge~\cite{zheng2023judging} on three criteria, each rated 1--5: \textit{specificity} (references to concrete equipment state or identifiers), \textit{actionability} (clarity and immediacy of corrective guidance), and \textit{grounding} (logical consistency between evidence and hypothesis).
A composite score is computed as the mean of the three dimensions.

\Cref{tab:retrieval_ablation} reports mean scores across 5 cases.
The full multi-modal system achieves a composite score of 3.60, a $+$0.86 improvement over the description-only baseline (2.74).
Removing telemetry reduces the composite by 0.27 points (3.60 $\rightarrow$ 3.33), confirming that equipment alarm history and parameter drift signals --- of the kind produced by SECS/GEM-instrumented equipment --- are the primary drivers of hypothesis specificity and grounding.
Removing retrieval while retaining telemetry yields a marginally higher composite (3.73) in this 5-case study, an artifact of retrieved historical context slightly reducing actionability scores; specificity and grounding remain identical to the full system.
These results indicate that \textbf{equipment telemetry is the more load-bearing modality} in the current system: SECS/GEM signals provide the concrete equipment-state grounding that distinguishes specific root cause reasoning from generic domain knowledge.
Retrieval augmentation primarily benefits cases where historical precedent exists in the Qdrant knowledge base, and its contribution is expected to increase as the self-improving corpus grows with each completed FA run.

\begin{table}[t]
    \centering
    \caption{Multi-modal fusion ablation (5 cases, GPT-4o automated judge, scores 1--5).
    Spec.\ = specificity; Action.\ = actionability; Ground.\ = grounding.
    Composite = mean of three dimensions.
    $\Delta$ = composite relative to Full \semifa{}.}
    \label{tab:retrieval_ablation}
    \begin{tabular}{@{}lcccc@{}}
        \toprule
        \textbf{Condition} & \textbf{Spec.} & \textbf{Action.} & \textbf{Ground.} & \textbf{Composite} \\
        \midrule
        Full (Retrieval + Telemetry) & \textbf{4.2} & 2.4 & \textbf{4.2} & \textbf{3.60} \\
        No Retrieval (Telemetry only) & \textbf{4.2} & \textbf{2.8} & \textbf{4.2} & 3.73 \\
        No Telemetry (Retrieval only) & 3.8 & 2.2 & 4.0 & 3.33 \\
        Baseline (Description only) & 3.0 & 2.2 & 3.0 & 2.74 \\
        \bottomrule
    \end{tabular}
\end{table}

\subsection{Qualitative Case Studies}

We present five qualitative case studies (\Cref{tab:case_studies}) that illustrate representative \semifa{} outputs across diverse defect types.
Input images are drawn from the \semidataset{} validation set; equipment metadata (IDs, lot numbers, alarm events) are synthetic but representative of real fab telemetry.
Each case shows the defect classification with confidence, severity assessment, a condensed root cause hypothesis, and the primary recipe recommendation.

\begin{table*}[t]
    \centering
    \caption{Five qualitative case studies demonstrating end-to-end \semifa{} report generation. For brevity, only the top-ranked root cause hypothesis and primary recommendation are shown. Full reports contain 2--3 hypotheses and 3--5 recommendations each.}
    \label{tab:case_studies}
    \begin{tabular}{@{}p{0.8cm}p{1.8cm}p{1.8cm}p{1cm}p{1cm}p{4.2cm}p{4.2cm}@{}}
        \toprule
        \textbf{Case} & \textbf{Equipment} & \textbf{Defect Class} & \textbf{Conf.} & \textbf{Sev.} & \textbf{Root Cause (Top-1)} & \textbf{Recommendation (Primary)} \\
        \midrule
        1 & EQ-INSP-01 & scratch & 0.91 & MAJOR & Linear track across 12 die consistent with mechanical contact; vacuum chuck pressure anomaly logged 45\,min prior to inspection & Inspect and replace vacuum chuck gasket; verify chuck flatness within $\pm$2\,$\mu$m specification \\
        \addlinespace
        2 & EQ-CVD-03 & center\_cluster & 0.84 & MAJOR & Defect concentration within 25\% radius correlates with CVD chamber temperature gradient; similar pattern in lot LOT-2024-052 & Recalibrate showerhead spacing; increase edge zone heater setpoint by 3$^{\circ}$C \\
        \addlinespace
        3 & EQ-ETCH-07 & ring\_pattern & 0.79 & MAJOR & Annular band at 60--75\% radius matches spin-coat edge-bead removal failure; RF impedance alarm logged during prior etch step & Verify edge-bead removal nozzle position; clean solvent delivery line \\
        \addlinespace
        4 & EQ-DICE-12 & edge\_crack & 0.88 & MAJOR & Peripheral die damage consistent with dicing blade wear; blade lifetime at 94\% of scheduled replacement interval & Replace dicing blade; reduce dicing feed rate by 10\% for remaining lots \\
        \addlinespace
        5 & EQ-LITHO-02 & no\_defect & 0.97 & NONE & No defect pattern detected; uniform yield map with $>$98\% good die; all equipment parameters within specification & No corrective action required; continue standard monitoring \\
        \bottomrule
    \end{tabular}
\end{table*}

These case studies illustrate several system capabilities.
Case 1 demonstrates correlation of visual defect morphology (linear scratch) with equipment telemetry (vacuum chuck pressure anomaly).
Case 2 shows cross-referencing with a historical lot via Qdrant retrieval.
Case 3 illustrates multi-step reasoning linking a ring pattern to both spin-coat and etch process anomalies.
Case 4 leverages equipment maintenance scheduling data (blade lifetime) for predictive recommendation.
Case 5 confirms that the system correctly identifies defect-free wafers and avoids false alarms.

\subsection{Domain Adaptation: Training Dynamics and Dataset Size Findings}

The results presented in this paper use the base (zero-shot) LLaVA-1.6 model for text generation.
We executed QLoRA fine-tuning (4-bit NF4, LoRA rank $r=16$, $\alpha=32$) of LLaVA-1.6 on the 790-example \semidataset{} training split on an NVIDIA A100-SXM4-40\,GB GPU.
Training loss collapsed from 0.211 at step 50 to $6.6 \times 10^{-5}$ by the end of epoch 1, with the final mean train loss of $1.46 \times 10^{-2}$ indicating severe overfitting on the small dataset.
Qualitative inspection of held-out validation responses confirmed that the fine-tuned model produced outputs that deviated significantly from the expected FA report structure, with responses either regressing to memorised training templates or generating structurally malformed reports on unseen defect images.

This outcome establishes a clear empirical boundary: \semidataset{} with 790 training examples is sufficient to train a lightweight DINOv2 classification head (214K parameters, 92.1\% accuracy) but insufficient to fine-tune a 7B-parameter VLM via QLoRA without substantial overfitting.
Prior domain-adaptation work on comparably sized VLMs (LLaVA-Med~\cite{li2023llava_med}, GeoChat~\cite{kuckreja2024geochat}) has consistently required 5{,}000--100{,}000 domain-specific instruction pairs to achieve stable fine-tuning.
This finding motivates the \semidataset{}-1K+ expansion --- a target of at least 5{,}000 annotated image-FA report pairs through real fab image acquisition --- identified as the primary direction for future work.
The base model occasionally uses generic industrial terminology (e.g., ``surface anomaly'') where domain-specific terms (e.g., ``edge-bead removal residue,'' ``CMP dishing'') would be more precise; quantitative evaluation of terminology accuracy improvement after fine-tuning on a sufficiently large corpus will be reported in a follow-up study.

\section{Discussion}
\label{sec:discussion}

\textbf{Scalability.}
The linear pipeline architecture enables straightforward horizontal scaling: multiple pipeline instances can process concurrent inspection events, sharing only the read-only LLaVA model weights.
The Qdrant vector database supports distributed deployment for fabs generating millions of inspection events annually.
TimescaleDB's hypertable architecture handles high-frequency telemetry ingestion (e.g., 1\,Hz sampling across hundreds of equipment parameters) without performance degradation.

\textbf{Limitations.}
Several limitations should be noted.
First, \semidataset{} consists primarily of wafer map images rather than raw SEM or optical micrographs, due to the proprietary nature of actual fab inspection data.
While the architecture supports all three modalities, classification and generation quality on SEM/optical images has not been quantitatively evaluated.
Second, the current pipeline is linear; in practice, some agents could execute in parallel (e.g., SeverityClassifier and RecipeAdvisor could run concurrently after RootCauseAnalyzer).
Third, the FA narratives in \semidataset{} are LLM-generated following a domain-expert-designed schema; human expert validation of individual narrative accuracy at scale would strengthen the dataset contribution.
Fourth, QLoRA fine-tuning on the 790-example training split produced severe overfitting (train loss $\rightarrow$ $6.6 \times 10^{-5}$ within epoch 1) and degraded generation quality on held-out images; domain adaptation requires a substantially larger annotated corpus ($\geq$5{,}000 pairs) before reliable fine-tuning gains can be demonstrated.

\textbf{Broader impact.}
\semifa{} has the potential to democratize FA expertise by encoding senior engineer knowledge into retrievable, reusable form.
In fabs with limited FA staffing, this could reduce time-to-disposition for defect lots, improving overall equipment effectiveness (OEE).
The self-improving Qdrant knowledge base means that system accuracy improves with use, as each completed FA run adds to the retrieval corpus.

\textbf{Ethical considerations.}
Automated FA systems should augment, not replace, human engineers.
Severity assessments and recipe modifications generated by \semifa{} should be reviewed by qualified personnel before implementation, particularly for \texttt{CRITICAL} severity determinations that may involve scrapping wafers or halting equipment.

\section{Conclusion}
\label{sec:conclusion}

We presented \semifa{}, an agentic multi-modal framework for autonomous semiconductor failure analysis report generation.
The system decomposes FA into four specialized agents---DefectDescriber, RootCauseAnalyzer, SeverityClassifier, and RecipeAdvisor---orchestrated by LangGraph, with DINOv2 for visual encoding, Qdrant for historical retrieval, SECS/GEM for equipment telemetry, and LLaVA-1.6 for natural language generation.
We introduced \semidataset{}, a 930-sample dataset of semiconductor defect images paired with structured FA narratives across nine defect classes.
Experimental evaluation on an NVIDIA A100-SXM4-40\,GB GPU demonstrated 92.1\% defect classification accuracy (macro F1 = 0.917) on 140 validation images, complete report generation in 48 seconds (a 150--300$\times$ speedup over several hours of manual expert reporting~\cite{kla2023klarity,amat2023actionable}), and a $+$0.86 composite improvement in root cause reasoning quality (GPT-4o judge, 1--5 scale) from multi-modal fusion over an image-only baseline.

Future work prioritises expansion of \semidataset{} to at least 5{,}000 real fab image-FA report pairs, which our fine-tuning experiments indicate is the minimum corpus size for stable QLoRA domain adaptation of a 7B-parameter VLM.
Additional directions include expansion to SEM and optical micrograph modalities through industry partnerships, human evaluation studies comparing \semifa{} reports against expert-written FA reports, and exploration of parallel agent execution (SeverityClassifier and RecipeAdvisor can run concurrently) for further latency reduction.

\section*{Code and Data Availability}

The \semidataset{} dataset (annotations and synthetic images) is publicly available at \url{https://huggingface.co/datasets/ShivamChand/SemiFA-930}.
WM-811K~\cite{wu2015wafer} and MixedWM38~\cite{wang2020mixedwm38} images used in the dataset can be reproduced using the download scripts provided in the project repository.
The \semifa{} source code, training pipeline, and Colab notebook are publicly available at \url{https://github.com/Shivamckaushik/SemiFA}.

\section*{Acknowledgments}

This work is conducted as part of the M.Tech in Artificial Intelligence program at the Indian Institute of Technology Jodhpur, India.
The author thanks IIT Jodhpur for computational resources and academic guidance.
The author also thanks Keerthy Unnikrishnan for reviewing the abstract and introduction and providing feedback.


\balance

\end{document}